%% file: main.tex
\documentclass[10pt,twocolumn,letterpaper]{article}

\usepackage[pagenumbers]{cvpr} % To force page numbers, e.g. for an arXiv version

\usepackage[accsupp]{axessibility}
\usepackage{graphicx}
\usepackage{amsmath}
\usepackage{amssymb}
\usepackage{booktabs}
\usepackage{mathtools}
\usepackage{multirow}
\usepackage[normalem]{ulem}
\usepackage{makecell}
\usepackage{float}
\usepackage{xspace}
\usepackage{wrapfig}

\usepackage[pagebackref,breaklinks,colorlinks]{hyperref}

\usepackage[capitalize]{cleveref}
\crefname{section}{Sec.}{Secs.}
\Crefname{section}{Section}{Sections}
\Crefname{table}{Table}{Tables}
\crefname{table}{Tab.}{Tabs.}

\def\cvprPaperID{1445} % *** Enter the CVPR Paper ID here
\def\confName{CVPR}
\def\confYear{2022}
\def\name{LASER\xspace}

\begin{document}

\title{LASER: LAtent SpacE Rendering for 2D Visual Localization}
%\title{Rendlas: Rendering in Latent Space - Visual Localization in 2D Floor Maps}
\author{Zhixiang Min$^1$ \quad Naji Khosravan$^2$ \quad Zachary Bessinger$^2$ \quad Manjunath Narayana$^2$ \\
Sing Bing Kang$^2$ \quad Enrique Dunn$^1$ \quad Ivaylo Boyadzhiev$^2$  \vspace{+0.3em} \\
$^1$Stevens Institute of Technology \qquad $^2$Zillow Group \vspace{-0em} \\
%{\tt\small \{zmin1,edunn\}@stevens.edu \{najik,zacharybe,manjun,singbingk,ivaylob\}@zillowgroup.com}
}
\maketitle

%%%%%%%%% ABSTRACT
\begin{abstract}
\vspace{-0.3em}
We present \name, an image-based Monte Carlo Localization (MCL) framework for 2D floor maps. 
\name introduces the concept of latent space rendering, where 2D pose hypotheses on the floor map are directly rendered into a geometrically-structured latent space by aggregating  viewing ray features. Through a tightly coupled rendering codebook scheme, the viewing ray features are dynamically determined at rendering-time based on their geometries (i.e. length, incident-angle), endowing our representation with view-dependent fine-grain variability.
Our codebook scheme effectively disentangles feature encoding from rendering, allowing the latent space rendering to run at speeds above 10KHz. 
Moreover, through metric learning, our geometrically-structured latent space is common to both pose hypotheses and query images with arbitrary field of views.
As a result, \name achieves state-of-the-art performance on large-scale indoor localization datasets (i.e. ZInD \cite{cruz2021zillow} and Structured3D \cite{zheng2020structured3d}) for both panorama and perspective image queries, while significantly outperforming existing learning-based methods in speed.

%We present \name, an image-based Monte Carlo Localization (MCL) framework for indoor environments with known 2D floor maps. 
%Our approach introduces the concept of latent space rendering,
%where learned environmental features are selected and rendered into a geometrically-structured (i.e. omnidirectional and rotationally-covariant) latent space representation.
%Importantly, through the use of metric learning, our geometrically-structured latent space is common to both input query images and candidate pose hypotheses.
%By decoupling environmental feature encoding from rendering, \name  achieves sampling throughput of over 10KHz.
%Moreover, based on the concept of rendering codebooks, our proposed circular feature is endowed with both global scope and view-dependent fine-grain variability. This enables state-of-the-art performance on large-scale indoor localization datasets (e.g. ZInD \cite{cruz2021zillow} and Structured3D \cite{zheng2020structured3d}), 
%while significantly outperforming existing learning-based methods in speed.
%while running significantly faster than existing learning-based methods.

\end{abstract}

%%%%%%%%% BODY TEXT
\vspace{-0.8em}
\section{Introduction}
\vspace{-0.1em}
Camera localization aims to estimate the spatial relationship between a given input image w.r.t. an environmental representation. 
Diverse application-driven variants have been addressed in the computer vision, robotics, and AR/VR literature. Particular problem instances are defined in terms of the scope of the pose geometric model (e.g. SE(2) vs SE(3)), the type of input query imagery (e.g. RGB, depth), as well as the type of the environmental geometric reference (e.g. geometric maps, registered image collections). Instances where the queries and the geometric reference share the same domain define camera localization as a direct geometric registration problem (e.g. ICP \cite{park2017colored}, SfM-based geometric verification \cite{schonberger2016structure}). Conversely, whenever query observations and the geometric reference are from different domains, it requires the design of integrative cross-modality data representations able to distinguish and associate input observations and reference data. 
%integrative cross-modality data representations are required to distinguish and associate input observations with the reference data.

\begin{figure}[t!]
    \centering
    \includegraphics[width=\columnwidth]{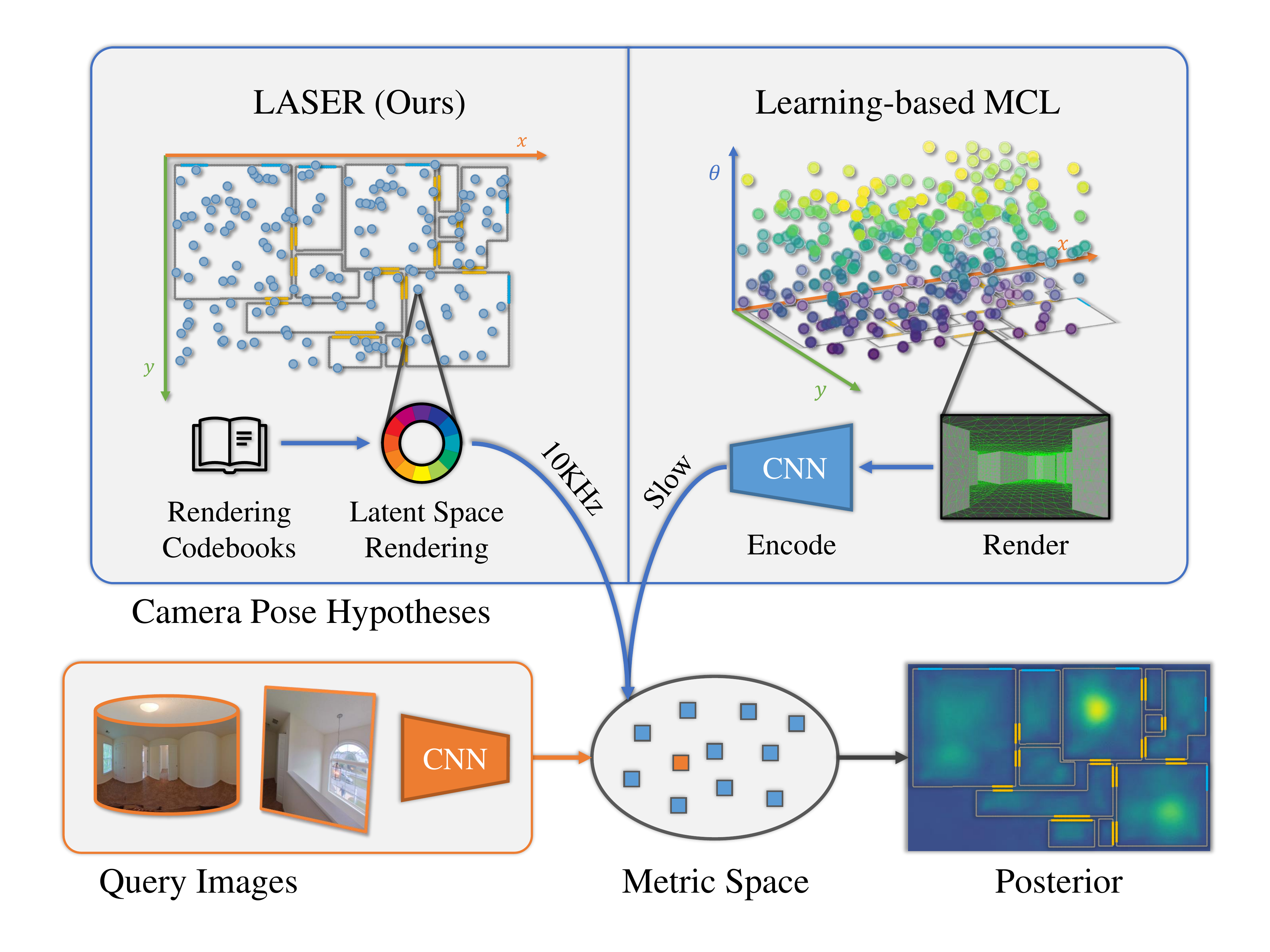}
    \vspace{-2em}
    \caption{\textbf{\name diagram.} Learning-based MCL frameworks encode camera pose hypotheses and query images into a common metric space
    to measure their similarities. 
    %where their similarities are measured. 
    Compared to existing works, \name directly renders latent features and has a reduced sampling dimension. }
    \vspace{-1.0em}
    \label{fig:teaser}
\end{figure}

This work focuses on solving for the camera pose of a query panorama/perspective image w.r.t. a 2D floor map, under the Monte Carlo Localization (MCL) framework \cite{dellaert1999monte}. 
MCL adopts a generative framework, where the solution (i.e. camera pose) space is systematically sampled to render observation hypotheses and states the problem in terms of a maximum likelihood search and/or optimization w.r.t. a query observation. 
Note that in this work, we solely focus on improving the measurement model, hence we are only interested in localizing individual queries without initialization. Yet, it is straight-forward to integrate our work into a full MCL framework with customizable temporal updates.
Conventional MCL methods \cite{dellaert1999monte, winterhalter2015accurate} require depth sensors and have limited robustness to environmental variability (e.g. furniture/object changes) due to their explicit geometric modeling.  Extensions leveraging image-based room layout estimation \cite{boniardi2019robot, wang2015lost} address content variability at the expense of imposing  environmental or capture assumptions such as Manhattan world, known ceiling or camera height. 
Recent supervised learning approaches \cite{howard2021lalaloc, karkus2018particle} have learned a common latent space for both synthesized and query observations. 
However, their explicit high-fidelity rendering and CNN-based encoding corresponding to individual pose hypothesis is computationally burdensome.
%However, the explicit rendering of high-fidelity observations corresponding to individual pose hypothesis is both computationally burdensome and semantically uninformative (i.e. layout depth).  
Given the time-sensitivity and the accuracy dependency on the number of samples for MCL applications,  their computational burden compromises estimation accuracy for online operation. Moreover, the attained latent space representations are not geometrically interpretable while lacking expressiveness and level of detail due to the coarse-level homogenization common to convolutional architectures.  
We address these challenges within a geometrically-structured metric learning framework, which performs latent-space rendering while prioritizing applicability to unseen environments, computational efficiency, estimation accuracy and robustness.

We circumvent expensive explicit rendering-and-encoding of sampling observations by directly rendering features in a learnable common metric space from a rasterized 2D floor map. 
Such latent space rendering is enabled by a rendering codebook, which allows map points to have view-dependent dynamic features for representing rendering-time dynamics (i.e. viewing ray geometries such as length and incident-angle).
Importantly, we structure said latent space to be geometrically meaningful by encoding visibility-based omni-directional observations into discretized circular (i.e. angular cyclical) representations.
%aggregating directional environmental feature information.
This representation, namely \emph{circular feature}, 
along with the view-dependent feature encoding from the rendering codebook, provides fine-grain structured descriptors for geometry and semantics at a high sampling FPS of 10KHz. 
Extensive experiments on Structured3D \cite{zheng2020structured3d} and ZInD datasets \cite{cruz2021zillow} show that our proposed framework significantly outperforms state-of-the-art frameworks both in accuracy and speed. The main technical contributions and innovations driving these performance gains are:

\noindent {\bf (1) Map-aware 2D visual localization framework}: While existing MCL frameworks render hypotheses over a local scope (i.e. contents visible to the camera), \name uses a 2D variant of the PointNet \cite{qi2017pointnet} to attain latent encoding from 2D point cloud maps. The PointNet learns global context from the map and provides the latent feature with map-level scope which improves \name's recall.

\noindent {\bf (2) Latent space rendering based on codebook scheme}: 
\name obviates the redundant rendering-and-encoding of an intermediate representation for individual samples by directly render features in the latent space. Powered by our rendering codebook scheme, the features are dynamically determined in rendering-time to encode fine-grain ray geometries. Such design achieves significantly higher sampling speed and accuracy at the same time.
%Powered by an adaptive rendering codebook that endows our previously encoded map point features with fine-grain rendering dynamics, our proposed latent space rendering directly renders features in a learnable common metric space. Obviating repeated and expensive rendering-and-encoding of a redundant intermediate representation for individual samples, \name achieves significantly higher sampling speed.

\noindent {\bf (3) Geometrically-structured metric learning}: 
\name structures the metric learning to be geometrically meaningful using a rotationally-covariant 2D omni-directional circular feature. 
Its fine-grain structured variability implicitly expresses environmental layouts for high-accuracy localization, and seamlessly supports query images with arbitrary field of views.
In addition, this choice implicitly encodes many orientations that reduces the rotation dimension from the MCL sampling space.

\begin{figure*}[t]
    \centering
    \includegraphics[width=2.1\columnwidth]{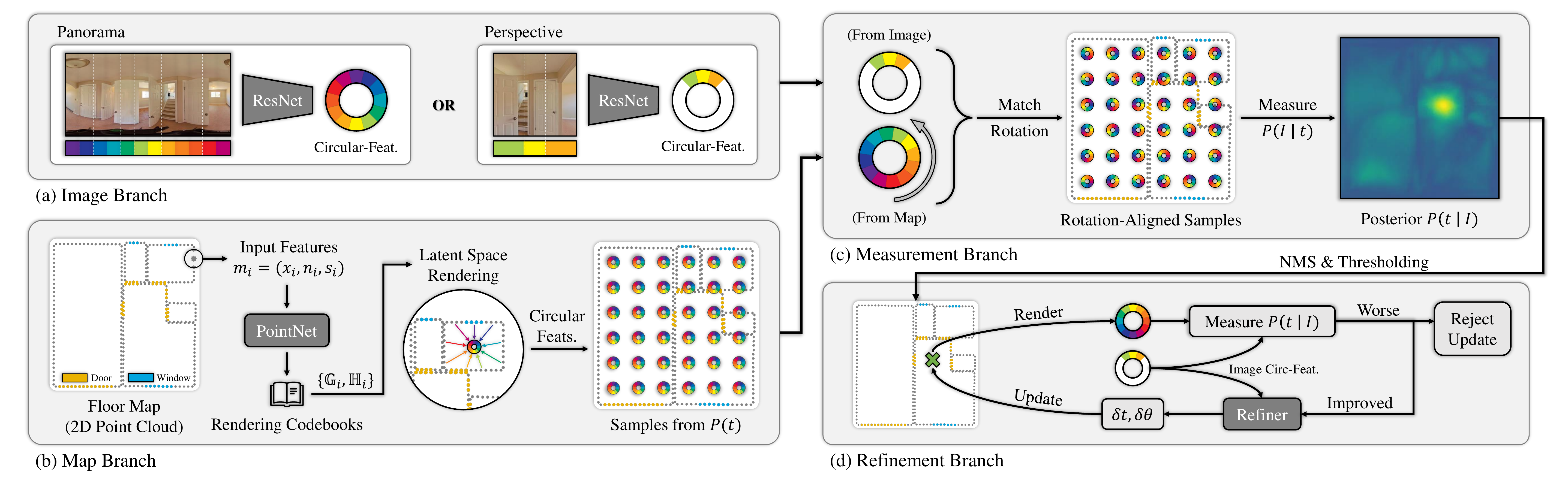}
    \vspace{-1.5em}
    \caption{\textbf{\name inference workflow.} Learnable components are shown in dark-gray boxes. (a) Panorama or perspective images are processed into circular features with ResNet50. (b) Rasterized 2D point cloud map is encoded using PointNet into rendering codebooks, from which our uniformly sampled circular features are rendered. (c) The similarities between image and map circular features are measured at their best matched rotation.  (d) The final estimation is iteratively refined until its likelihood stops improving. 
    }
    \vspace{-0.5em}
    \label{fig:workflow}
\end{figure*}

%\vspace{-0.5em}
\section{Related Works}
The general 6-DoF relocalization methods either explicitly \cite{sarlin2021back, sarlin2019coarse, brachmann2017dsac, sarlin2020superglue, sun2021loftr, zhou2021patch2pix, yi2018learning, taira2018inloc, yang2019sanet, min2021voldor+} or implicitly \cite{kendall2015posenet, balntas2018relocnet, ding2019camnet} find appearance correspondences between the query image and scene representations (e.g. images with known camera poses, sparse/dense 3D reconstructions). The camera pose can then be predicted from neural networks \cite{kendall2015posenet, balntas2018relocnet, ding2019camnet} or can be recovered using SfM methods \cite{zheng2015structure, schonberger2016structure, Zhou_2018_ECCV, tang2018ba, teed2018deepv2d, wei2020deepsfm, min2020voldor} from explicit correspondences. The appearance dependency limits their robustness to appearance changes and cannot work with pure geometry maps (e.g. occupancy map). 

Monte Carlo Localization (MCL) \cite{dellaert1999monte, winterhalter2015accurate, mendez2018sedar, chu2015you} is the most popular framework for 2D localization on pure geometry maps. With our interest solely on single query, MCL defines a measurement model over geometry observations from depth sensors and compares it with simulated observations sampled from the floor map. The methods limit the input type to geometric measurements, which also limit their robustness to geometry variation/occlusions in the map (e.g. changes in furniture/object). Some extensions of MCL take intensity images as input. Boniardi \etal \cite{boniardi2019robot} recover room geometry by explicitly extracting room layout edges using CNNs. Wang \etal \cite{wang2015lost} further cooperates semantic information for localization in large indoor spaces. All these methods are subject to strong assumptions such as Manhattan world and known ceiling or camera height, which limit their applications.

Recent learning-based methods \cite{howard2021lalaloc, karkus2018particle, jonschkowski2018differentiable} extend the MCL into a metric learning framework. They utilize learnable CNNs to encode query images and render location into the same metric latent space to estimate their similarity. For rendering a given camera pose, PfNet \cite{karkus2018particle} uses a spatially transformed bird's-eye map image, while LaLaLoc \cite{howard2021lalaloc} assumes known camera and ceiling height, and renders the layout depth image. Given the heavy rendering-and-encoding process, these methods are computationally burdensome, which limits their performance for time-sensitive or SWaP-constrained applications.

Neural Radiance Field (NeRF) \cite{mildenhall2020nerf, martin2021nerf} is a newly emerging area that synthesizes photo-realistic images from scene-specific neural representations learned using back-propagation. In contrast, our scene representation is the rendering codebooks inferred from PointNet, whose estimation process is scene-agnostic. Moreover, our latent space rendering synthesizes view-dependent latent features.

\section{Method}
%This section details our framework. It starts with the problem formulation, followed by the three main modules (i.e. map branch, image branch and, refinement branch), and is concluded with the training and inference strategies. 
% Problem Formulation
\subsection{Problem Formulation}
We define camera localization as estimating the 2D pose $\mathbf{p}^*  \!=\! [\mathbf{t},\theta] \!\in\! SE(2)$, associated with a given query image $\mathbf{I}$, w.r.t. a reference map $\mathbf{M}$. 
The pose parameters $\mathbf{t} \!\in\! \mathbb{R}^2$ and $\theta \!\in\! [0,2\pi)$ define, respectively, the camera's planar displacement vector and yaw axis rotation. The input query  $\mathbf{I}$ may be panorama or a perspective image with known FoV. We do not limit the format of map $\mathbf{M}$, but assume it encodes the occupancy information within a 2D plane.

% Monte Carlo Localization.
\noindent\textbf{Monte Carlo Localization.} The general Monte Carlo Localization (MCL) framework \cite{dellaert1999monte} defines a measurement model $P(\mathbf{I}\,|\, \mathbf{p}; \mathbf{M})$, which expresses the likelihood of image $\mathbf{I}$ being observed at camera pose $\mathbf{p}$ on map $\mathbf{M}$. Henceforth, we obviate the non-random parameter $\mathbf{M}$ from our formulation for simplicity. The posterior distribution of 
%camera pose 
$\mathbf{p}$ after observing  $\mathbf{I}$ is the solution of interest. Following Bayes rule, MCL estimates the posterior distribution $P(\mathbf{p} \,|\, \mathbf{I})$ as
\begin{equation}\label{eq:framework_bayes}
    P(\mathbf{p} \,|\, \mathbf{I}) = \frac{P(\mathbf{I}\,|\, \mathbf{p}) P(\mathbf{p})}{P(\mathbf{I})}
\end{equation}
where $P(\mathbf{I})$ is a normalization constant that can be safely ignored, while $P(\mathbf{p})$ is the prior camera pose distribution, which we assume uniformly distributed within the map area.
Finally, the full posterior can be approximated by drawing particles from $P(\mathbf{p})$ whose likelihoods will be estimated using the measurement model as defined in Eq.\ref{eq:measurement_model_full}.

% Localization as Metric Learning.
\noindent\textbf{Localization as Metric Learning.} The measurement model $P(\mathbf{I}\,|\, \mathbf{p})$ within the MCL framework defines the similarity across the camera pose and image domains. We adopt deep metric learning \cite{hoffer2015deep} to learn a unified metric space for comparing cross-domain similarity between the query image and camera pose hypotheses as shown in Fig.\ref{fig:workflow}. We detail how images and camera poses are encoded in to the metric space in \S\ref{sec:image_branch} and \S\ref{sec:floorplan_branch}, respectively. 

% Circular Feature
\noindent \textbf{Circular Feature.} Contrary to conventional flattened descriptors used in metric learning, we introduce circular features to encode spatial visibility, leading to our geometrically-structured metric learning. We define a circular feature as an ordered set of feature vectors 
\begin{equation} \label{eq:circular_feature}
    \mathbb{F}=\{\mathbf{f}^\alpha \,|\, \alpha=0 \cdots V\!-\!1\}
\end{equation}
where $V$ is the number of feature segments. Each feature segment $\mathbf{f}^\alpha \!\in\! \mathbb{R}^D$ encodes a local directional FoV of $\frac{2\pi}{V}$ radians in the range $\big[ \frac{2\pi\alpha}{V}, \frac{2\pi (\alpha\!+\!1)}{V} \big)$ on the 2D plane. We denote this ordered set $\mathbb{F}$ as a circular feature since the first and last feature segments correspond to adjacent FoVs. With this design, the omni-directional 2D spatial information is implicitly encoded in the order of feature segments.

% Measurement Model
\noindent\textbf{Measurement Model.}  We first define the similarity measurement between two circular features $\mathbb{F}_i=\{\mathbf{f}^\alpha_i \,|\, \alpha=0 \cdots V\!-\!1\}$ and $\mathbb{F}_j=\{\mathbf{f}^\alpha_j \,|\, \alpha=0 \cdots V\!-\!1\}$ as
\begin{equation} \label{eq:circular_feat_sim}
    \mathcal{S} \big( \mathbb{F}_i, \mathbb{F}_j \big) = \frac{\sum_{\alpha=1}^V cos(\mathbf{f}^\alpha_i, \mathbf{f}^\alpha_j)}{2V}+0.5 
\end{equation}
where $cos(\cdot, \cdot)$ computes the vector cosine similarity, and the function output is normalized to $[0,1]$. We further define a rotating operator $\mathcal{R} \big( \mathbb{F}, \theta \big)$ to rotate the underlying spatial information of a circular feature $\mathbb{F}$ with a given angle $\theta$ by
\begin{equation} \label{eq:rot_circular_feat}
    \mathcal{R} \big( \mathbb{F}, \theta \big) = \{\mathbf{f}^{ (\alpha+\frac{V\theta}{2\pi}) \% V  } \,|\, \alpha=0 \cdots V\!-\!1\} 
\end{equation}
where the  1D feature space is linearly interpolated when the indexing yields non integer values.
%(i.e. using the two nearest neighbors with indexes $\lfloor \alpha+\frac{V\theta}{2\pi} \rfloor \% V$ and $\lceil \alpha+\frac{V\theta}{2\pi} \rceil \% V$).
Finally, we define the measurement model as
\begin{equation} \label{eq:measurement_model_full}
    P(\mathbf{I}\,|\, \mathbf{p}) = P(\mathbf{I}\,|\, \mathbf{t}, \theta) = A \cdot \mathcal{S}\big(\mathbb{F}_{\mathbf{I}}, \mathcal{R} \big(\mathbb{F}_{\mathbf{t}},\theta \big) \big)
\end{equation}
where $A$ is the PDF normalization constant, and $\mathbb{F}_{\mathbf{I}}$ and $\mathbb{F}_{\mathbf{t}}$ are circular features encoded from the query image and rendered at location $\mathbf{t}$ on the map respectively. 

% Rotation Reduction
\noindent\textbf{Rotation Reduction.} 
As MCL needs a large number of samples to approximate the camera pose posterior in $SE(2)$, we systematically reduce the rotation dimension from the MCL sampling step.
For a sample location $\mathbf{t}$ with a canonical orientation, the optimal relative rotation of its circular feature $\mathbb{F}_\mathbf{t}$ w.r.t. an image circular feature $\mathbb{F}_\mathbf{I}$ can be found by
\begin{equation} \label{eq:optimal_rotation}
    \theta^{opm}_{\mathbf{t}} = \underset{\theta_{\mathbf{t}}}{argmax\;} \mathcal{S}\big(\mathbb{F}_{\mathbf{I}}, \mathcal{R}\big(\mathbb{F}_{\mathbf{t}},\theta_{\mathbf{t}}\big)\big)
\end{equation}
Substituting into Eq.\ref{eq:measurement_model_full}, we attain a simplified measurement model obviating $\theta$ and conditioned solely on $\mathbf{t}$ as
\begin{equation} \label{eq:measurement_model_partial}
    P(\mathbf{I}\,|\, \mathbf{t}) = A' \cdot \mathcal{S}\big(\mathbb{F}_{\mathbf{I}}, \mathcal{R}\big(\mathbb{F}_{\mathbf{t}},\theta^{opm}_{\mathbf{t}}\big)\big)
\end{equation}
For solving Eq.\ref{eq:optimal_rotation}, we rotate $\mathbb{F}_{\mathbf{t}}$ with uniformly sampled $\theta_{\mathbf{t}}$ in $[0,2\pi)$, and keep the best. 
%Although a clever trust-region optimizer can be used to refine the estimation, we found the naive sampling works very well even with very few samples (i.e. 8 samples) as shown in Fig.\ref{fig:perf_over_sampling_density}. 
This discretized search initializes rotation to a rough value, which will later be refined as in \S\ref{sec:refinement_branch}.
The rotation matching process is highly efficient since it
%is different from MCL sampling since it 
reuses the same circular feature and does not render new hypotheses, where its throughput is detailed in Table.\ref{table:timing}.

%Map branch
\subsection{Map Branch} \label{sec:floorplan_branch}
In this section, we show how circular features are rendered from 2D floor maps with a given camera pose.
%Map branch is responsible for representing and sampling floormaps in the latent space.

%Map branch - 2D Point Cloud Map
\noindent\textbf{Rasterized 2D Point Cloud Map.} Given a general 2D map representation $\mathbf{M}$ (e.g. floorplan or occupancy grid) encoding area occupancy info, we uniformly sample points on the occupancy boundaries (i.e. walls) to form a 2D point cloud $\mathbb{M}=\{ \mathbf{m}_i \,|\, i=0 \cdots N\!-\!1\}$. Each point $\mathbf{m}_i = [\mathbf{t}_i, \mathbf{n}_i, \mathbf{s}_i]$ encodes its location $\mathbf{t}_i$, normal vector $\mathbf{n}_i$ and  optional semantic information $\mathbf{s}_i$. Available semantic information (e.g. door or window labels), are encoded as multiple binary masks appended to the point representation. 

\begin{figure}[t]
    \centering
    \vspace{-0.5em}
    \includegraphics[width=0.9\columnwidth]{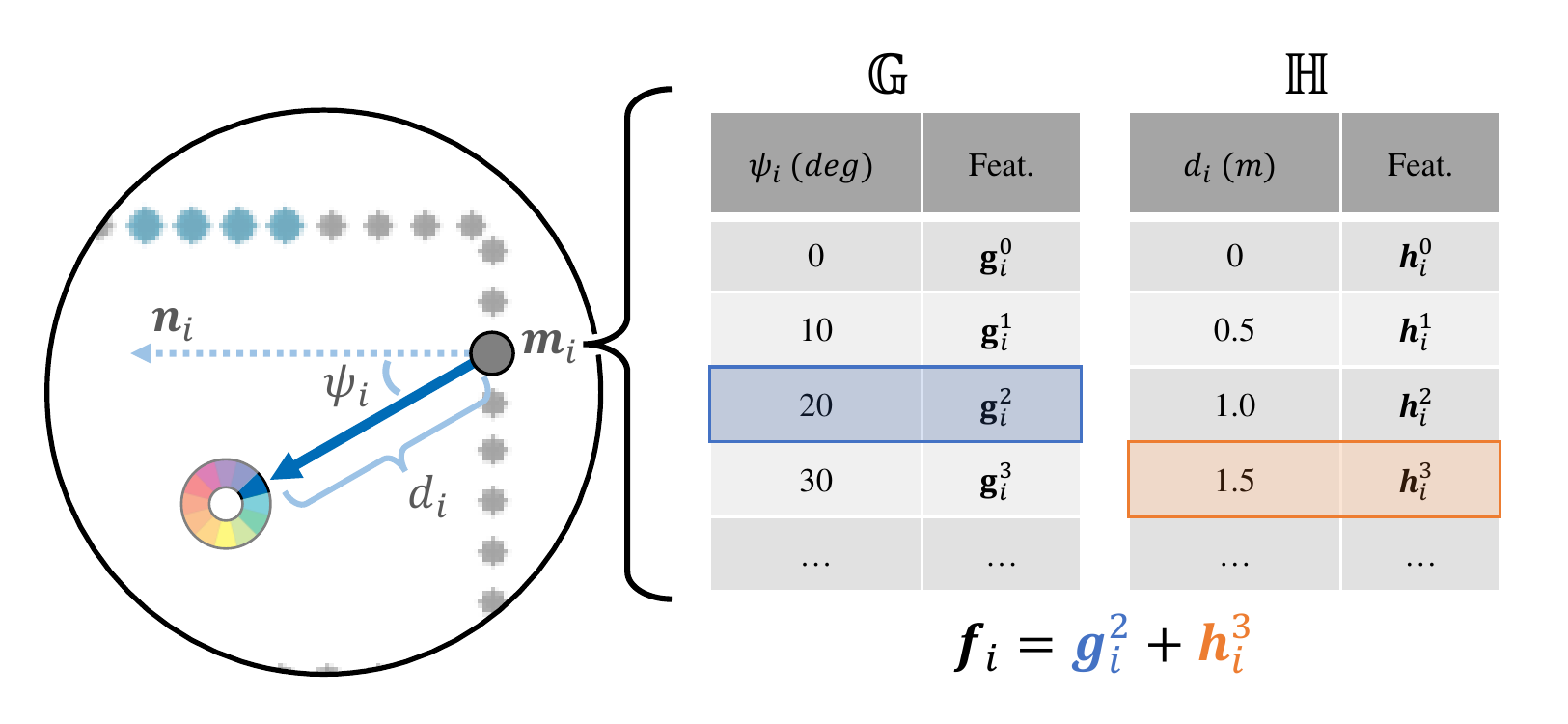}
    \caption{\textbf{Rendering codebook.} The map point features are dynamically (view-dependent) determined at rendering-time by the codebook and rendering dynamics. Linear interpolation is applied between adjacent codes in the codebook. Features from multiple codebooks are aggregated by summation.}
    \vspace{-0.5em}
    \label{fig:codebook_diagram}
\end{figure}

%Map branch - Latent Space Rendering
\noindent\textbf{Latent Space Rendering.} 
To circumvent the inefficient two-stage rendering-and-encoding process, we propose latent space rendering that directly renders circular features for given locations by aggregating 
%PointNet 
features from visible map points. 
However, visibility proved to be a necessary but insufficient cue for the effective selection and rendering of latent space features (as shown in Fig.\ref{fig:scoremaps_pano}). 
More specifically, visibility for static environments is locally constant at most sampling locations, providing a limited spatial context. To mitigate potential homogenization of our representations, we analyze fine-grain rendering dynamics such as length and incident-angle of the viewing rays between features and sampling location, and define an adaptive rendering mechanism. 
%However, naively doing so produces very poor results as shown in Fig.\ref{fig:scoremaps_pano} since it only aggregates the visibility information and disregards the rendering-time dynamics such as length and incident-angle of the viewing-rays. The rendering dynamics encodes important observation geometry information that MCL framework use for localization.
%However, naively aggregating feature vectors from visible map points to the rendering location does not represent the location satisfyingly as shown in Fig.\ref{fig:scoremaps_pano}. This is because aggregating fixed feature vectors from map points only provides visibility information while discards rendering-time dynamics such as the length or incident-angle of the viewing-ray, which are indispensable geometry information for localization.
\\ \noindent\textbf{Rendering Codebooks}. 
We propose an over-specified latent space in order to endow map points with view-dependent features to encode rendering dynamics. We encode the 2D point cloud map with a 2D variant of PointNet \cite{qi2017pointnet} to assign each map point $\mathbf{m}_i$ two sets of features $\mathbb{G}_i=\{ \mathbf{g}^\beta_i \,|\, \beta=0 \cdots G\!-\!1\}$ and $\mathbb{H}_i=\{ \mathbf{h}^\gamma_i \,|\, \gamma=0 \cdots H\!-\!1\}$,  denoted as distance and incident-angle codebook respectively. Features in the codebook have the same dimension as circular feature segments $\mathbf{g}^\beta, \mathbf{h}^\gamma \!\in\! \mathbb{R}^D$. At rendering-time, map point features are chosen from the codebooks based on their distance and incident-angle w.r.t. the rendering location as shown in Fig.\ref{fig:codebook_diagram}. Formally, assume a rendering location $\mathbf{\hat{t}}$ and a map point $\mathbf{m}_i=[\mathbf{t}_i, \mathbf{n}_i, \mathbf{s}_i]$, let $\mathbf{d}_i=\mathbf{t}_i - \mathbf{\hat{t}}$, we can compute its rendering dynamics by
\begin{gather} 
    d_i = \|\, \mathbf{d}_i \,\| \\
    \psi_i = atan2 \big( \| \mathbf{d}_i \times \mathbf{n}_i \| ,\, \mathbf{d}_i \!\cdot\! \mathbf{n}_i \big)
\end{gather} 
where $d_i$ and $\psi_i$ are distance and incident-angle respectively. The clockwise incident-angle $\psi \!\in\! [0,2\pi)$  distinguishes the four quadrants. With $\mathbf{m}_i$'s associated codebooks $\mathbb{G}_i$ and $\mathbb{H}_i$, its feature $\mathbf{f}_i$ is then determined by 
\begin{equation}
    \mathbf{f}_i = \mathbf{g}_i^{\frac{G \psi_i}{2\pi}} + \mathbf{h}_i^{ min(\frac{Hd_i}{d_{max}}, H)}
\end{equation}
where $d_{max}$ is a pre-defined maximum distance for the distance codebook. Similar to Eq.\ref{eq:rot_circular_feat}, for non-integer indexing, we linearly interpolate between its two closest codes. Finally, if $\mathbf{m}_i$ passes the visibility test to location $\mathbf{\hat{t}}$, we project $\mathbf{f}_i$ to circular feature $\mathbb{F}_{\mathbf{\hat{t}}}=\{ \mathbf{f}^\alpha_{\mathbf{\hat{t}}} \,|\, \alpha=0 \cdots V\!-\!1\}$ by
\begin{equation}
    \mathbf{f}^{\frac{V\omega_i}{2\pi}}_{\mathbf{\hat{t}}} = \mathbf{f}_i
\end{equation}
where $\omega_i$ is the angle of viewing ray $\mathbf{d}_i$. Finally, the projected map point features are averaged into each segment. See supplementary for more rendering details.

%Image branch
\subsection{Image Branch} \label{sec:image_branch}
In this section, we show how circular features are extracted from panorama and perspective images.
%Image branch extract features and represents panorama or perspective images with circular features.

%Image branch - Circular Feature from Panorama
\noindent\textbf{Circular Feature from Panoramas.} For panorama images in equirectangular projection,  each image column 
%in the equirectangular image 
corresponds to a fixed horizontal FoV, as shown in Fig.\ref{fig:workflow}(b). 
Such capture configuration facilitates a bijective mapping between our groups of adjacent input image columns and the segments in our rendered circular representation. The query panorama is fed into a ResNet50 \cite{he2016deep} encoder to obtain a feature map, which is subsequently squeezed by averaged-pooling in the vertical dimension to comply with the feature dimensions of our circular segments, and averaged-pooled again in the horizontal direction to $V$ elements, in accordance to the preconfigured number of feature segments in each circular feature.

%Image branch - Circular Feature from Perspective image
\noindent\textbf{Circular Feature from Perspective Images.}  We assume input perspective query images having known FoV and zero pitch/roll angle w.r.t. the ground plane. Note that for indoor query images,  pitch/roll angles may be rectified from vanishing point estimates \cite{guillou2000using, davidson2020360}. Accordingly, each perspective image column  corresponds to a non-fixed but known horizontal FoV as shown in Fig.\ref{fig:workflow}(a). We extract an image feature map with a ResNet50 encoder, using average pooling to squeeze the vertical dimension, and apply a perspective-to-equirectangular transform on the feature map to get the final circular feature. Since perspective images have no more than 180° FoV, its circular feature will have segments without assigned values, which will be masked out in the computation. Eq.\ref{eq:circular_feat_sim} will also be re-normalized to have range $[0,1]$. In supplementary, our model's  robustness  to capture pitch/roll angle alignment noise is detailed.

%Refinement Branch
\subsection{Refinement Branch} \label{sec:refinement_branch}
%To avoid inefficient exhaustive sampling for the MCL methods to achieve better accuracy,
We address the discretized pose sampling nature of our MCL approach, by proposing a light-weight continuous refinement branch to improve upon the current estimation.
As  Fig.\ref{fig:workflow}(d) shows, with current best estimation $\mathbf{t}^*$ and $\theta^*$, our refinement branch takes two circular features $\mathbb{F}_{\mathbf{I}}$ and $\mathcal{R}\big(\mathbb{F}_{\mathbf{t}^*}, \theta^*\big)$ as input. The refinement network uses two 1D convolution layers with circular padding followed by a fully-connected layer to predict two offsets $\delta\mathbf{t}$, $\delta\theta$ for translation and rotation respectively. Then we render the updated map circular feature $\mathcal{R}\big(\mathbb{F}_{\mathbf{t}^*\!+\!\delta\mathbf{t}}, \theta^*\!+\!\delta\theta\big)$ and compute its similarity to $\mathbb{F}_{\mathbf{I}}$ using Eq.\ref{eq:circular_feat_sim}. If the similarity score improved upon the original camera pose, we accept the step and iterate, otherwise we consider the refinement converged. The first refinement is always accepted to unquantize the estimation. This refinement usually converges within 3 iterations.

% Training & Inference
\subsection{Training \& Inference}
%In this section, we describe our training and inference process.
%The training and implementation details are as follows:

% Training & Inference - Triplet Loss
\noindent\textbf{Triplet Loss.}
We use triplet loss \cite{schroff2015facenet} to learn a mutual metric space between images and maps. To form a triplet, we let the image circular feature $\mathbb{F}_{\mathbf{I}}$ be the anchor, the map circular feature at ground truth camera pose $\mathbb{F}^{+} \!=\! \mathcal{R}\big(\mathbb{F}_{\mathbf{t}_{gt}}, \theta_{gt}\big)$ be the positive, and the map circular feature at a randomly sampled camera pose $\mathbb{F}^{-} \!=\! \mathcal{R}\big(\mathbb{F}_{\mathbf{t}_{rnd}}, \theta_{rnd}\big)$ to serve as the negative. Then the triplet loss is defined as
\vspace{-0.1em}
\begin{equation} \label{eq:triplet_loss}
    \mathcal{L}_{triplet} = 2 \cdot max
    \Big(  
    \mathcal{S} \big(\mathbb{F}_{\mathbf{I}}, \mathbb{F}^{+}  \big) - 
    \mathcal{S} \big(\mathbb{F}_{\mathbf{I}}, \mathbb{F}^{-}  \big)   \!+\! 0.5 ,\, 0
    \Big) 
\end{equation}

\begin{figure*}[h]
    \centering
    \includegraphics[width=2\columnwidth]{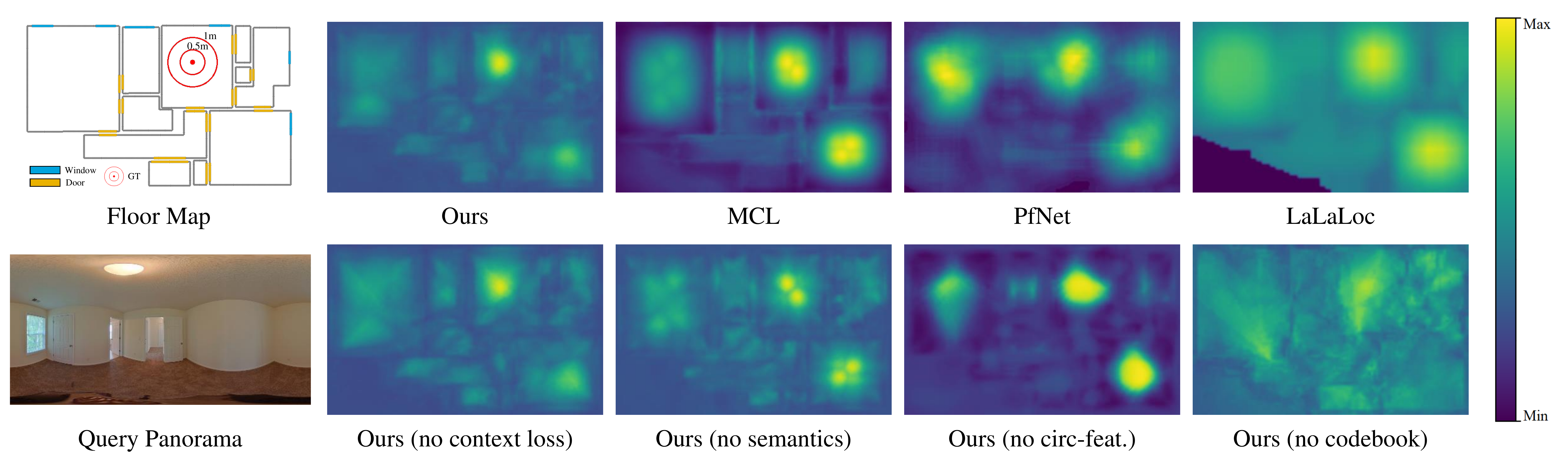}
    \vspace{-0.5em}
    \caption{\textbf{Posterior map comparison with panorama query.} Our method leverages semantic information and provides a single clear maximum without ambiguities compared to baselines. Without context loss, the posterior map becomes slightly nosier. Without semantic labels,  symmetric ambiguities emerge. Without circular feature, an accurate clear maximum cannot be identified from the clusters. Without codebook, the posterior map fails to show clear maxima. }
    \vspace{-0.5em}
    \label{fig:scoremaps_pano}
\end{figure*}

% Training & Inference - Context Loss
\noindent\textbf{Context Loss.} 
Our similarity function $\mathcal S$, and consequently also our triplet loss $\mathcal{L}_{triplet}$, relies on aggregating element-wise comparisons, which effectively disregarding any intra-feature context. 
We design an additional context loss, providing feature segments a wider scope of its circular feature for learning context information (i.e. properties of the room/map). We first define circular feature context $ \overline{\mathbb{F}}$ as the mean of its normalized feature segments
\begin{equation}
    \overline{\mathbb{F}} = \frac{\sum_{\alpha=1}^V \mathbf{f^\alpha} / \|\mathbf{f^\alpha}\| }{V}
\end{equation}
which we apply to our training triplet  similarly to Eq.\ref{eq:triplet_loss}
\begin{equation} \label{eq:context_loss}
    \mathcal{L}_{context} = max \big( cos( \overline{\mathbb{F}}_{\mathbf{I}}, \overline{\mathbb{F}}^{_+} ) \!-\! cos( \overline{\mathbb{F}}_{\mathbf{I}}, \overline{\mathbb{F}}^{_-}  ) \!+\! 1.0,\, 0 \big)
\end{equation}
With the context loss, circular features achieve better coarse level expressiveness, improving  recall for query images with limited FoV, as shown in Table.\ref{table:ablation}. This loss also acts as a regularizer by mitigating feature segments having large variance, leading to a smoother posterior estimation as shown in Fig.\ref{fig:scoremaps_pano}. 

% Training & Inference - Refinement Loss
\noindent\textbf{Refinement Loss.} For training the refinement branch, we sample circular features within a 0.5 meter radius and a 30 degree angle from the ground truth camera pose. We supervise the refinement branch using a regression loss as
\begin{equation} \label{eq:refine_loss}
\begin{aligned}
\mathcal{L}_{refine\_t} &= \big\| (\mathbf{t}_{gt} \!-\! \mathbf{t}^*) \!-\! \delta\mathbf{t} \big\|
\\
\mathcal{L}_{refine\_r} &= min \Big( \big|(\theta_{gt} \!-\! \theta^*) \!-\! \delta\theta \big| ,\,  2\pi \!-\! \big| (\theta_{gt} \!-\! \theta^*) \!-\!\delta\theta \big| \Big)
\end{aligned}
\end{equation}

% Training & Inference - Training Details
\noindent\textbf{Implementation Details.}
For triplet and context loss we sample 100 negative samples and broadcast the single ground truth (GT) sample for each training iteration. For refinement loss, we sample 20 hard negatives near the GT camera pose with a disturbance sampled from uniform distribution bounded in 30 degree and 0.5 meter radius. We combine the mean of all losses with equal weights. We set the hyper-parameters as $G\!=\!H\!=\!32$, $V\!=\!16$, $D\!=\!128$ and $d_{max}=10\,m$ consistently throughout the benchmarking. We sample the map into 2D point cloud with a 10 cm interval at occupancy boundaries. We render circular features for a $0.1m \times 0.1m$ uniform grid within the map range. For solving the relative rotations in Eq.\ref{eq:optimal_rotation}, we evaluate $16$ uniformly sampled angles and keep the best. Finally, the posterior distribution is estimated using Eq.\ref{eq:framework_bayes},\ref{eq:measurement_model_partial}. To extract final estimations from the posterior grid map, we apply a $3\!\times\!3$ non-maximum suppression to extract the maximums. For maximums that have larger score than a threshold (i.e. 0.8), we send them into the refinement branch to get the final estimations with their likelihoods as uncertainty estimation. Sorting by their likelihoods, a top-k estimation is available. More details see supplementary.

\section{Experiments}
\subsection{Datasets and Setups}
We test on Zillow Indoor Dataset (ZInD) \cite{cruz2021zillow} and Structured3D dataset (S3D) \cite{zheng2020structured3d}. ZInD  provides 1,575 real world unfurnished residential homes with 59,361 panoramas, while S3D is a synthetic dataset containing 3,500 houses created by designers with 21,835 panoramas each having different lighting and furnishing levels. Both datasets provide 2D floor maps with windows and doors labels, and the 360° panoramas are registered to the floor maps.

\begin{figure}[]
    \centering
    \includegraphics[width=0.95\columnwidth]{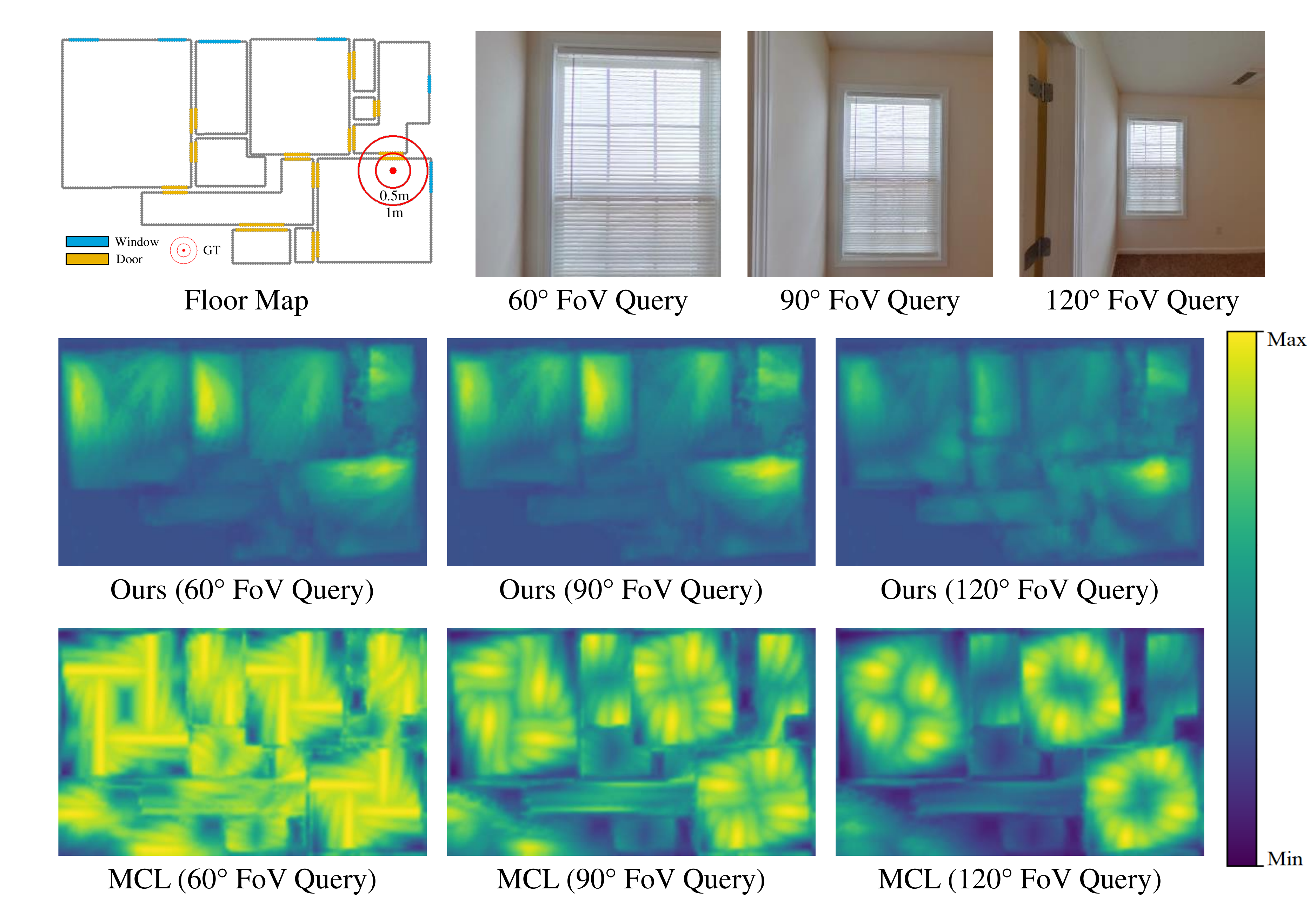}
    \vspace{-1em}
    \caption{\textbf{Posterior map comparison with perspective queries.} With increasing query FoV, our method has increasing confidence around the GT location. Compared to MCL, our method leverages semantic information, which greatly reduces ambiguities. }
    \vspace{-0.5em}
    \label{fig:scoremaps_pview}
\end{figure}

\begin{table*}[ht]
    \renewcommand{\arraystretch}{1.1}
    \centering
    \resizebox{2.05\columnwidth}{!}{%
        \input{tabels/table_baselines}
    }
    \caption{\textbf{Comparison with baselines on ZInD and S3D (fully furnished) dataset.} For indicating accuracy, we report median translation (terr) and rotation error (rerr) for all instances localized under 1m. We report recall at different translation accuracy levels, recall for inliers ($<$1m and $<$30°) and top-3 recall at 1m.}
    \vspace{-0.3em}
    \label{table:baselines}
\end{table*}

We follow the official training/testing splitting for both datasets, and train separate models for panorama and perspective images. 
For benchmarking (Table.\ref{table:baselines} and Fig.\ref{fig:perf_over_fov}), where we report performance for multiple perspective image FoVs, the model for perspective images is trained using images with FoVs uniformly sampled from 45° to 135°. For ablation study experiments, we report performance on a representative 90° FoV, where the model is trained exclusively on images with 90° FoV. All perspective images are cropped from panoramas in the dataset with zero pitch, roll angles, a random yaw angle and equal horizontal/vertical FoVs. 

% Benchmark
\subsection{Benchmark} \label{sec:benchmark}
We use LaLaLoc \cite{howard2021lalaloc}, PfNet\cite{karkus2018particle} and MCL \cite{dellaert1999monte} as our baseline methods
(see Table.\ref{table:baselines}). For MCL, we simulate a 72-ray 2D LiDAR giving ground-truth distance without noise as its input. The PfNet takes input a semantic labelled floor map same as the one in Fig.\ref{fig:scoremaps_pano}. For LaLaLoc, we follow their original protocol and take panoramas with known rotation as input and samples at 0.5m$\times$0.5m grid.
For PfNet and MCL, we sample a 0.1m$\times$0.1m grid with 16 rotation angles same to our framework. We also attached a random baseline to show the statistics of the dataset. 
To compare with methods not utilizing semantic map labels (i.e. MCL, LaLaLoc), we report our performance without using semantic information as well. \name exhibits superior accuracy and recall compared to our baseline methods. When the semantic labeling is not available, \name recall rate is close to MCL with the GT depths as input. 
For panorama queries, \name exhibits slightly better accuracy on S3D compared to ZInD due to the slight annotation errors in ZInD. For perspective query images, distance information becomes harder to extract since the room layout is sometimes not observable. In such case, \name works better with furnished rooms in S3D, where the furniture  provide cues for measuring distance. See supplementary for a more detailed discussion.

\begin{table}
    \renewcommand{\arraystretch}{1.05}
    \centering
    \resizebox{\columnwidth}{!}{%
        \input{tabels/table_ablation}
    }
    \vspace{-0.5em}
    \caption{\textbf{Ablation study over model components.} The base model uses the estimation from the posterior map without refinement. Without codebook, each map point is assigned a fixed feature. Without PointNet, all map points with the same semantic label share a fixed codebook. Without the circular structure (i.e. $V\!=\!1$), the feature becomes agnostic to rotation. }
    \vspace{-0.2em}
    \label{table:ablation}
\end{table}

% Performance Study
\subsection{Performance Studies}

\begin{figure}[]
    \centering
    \includegraphics[width=\columnwidth]{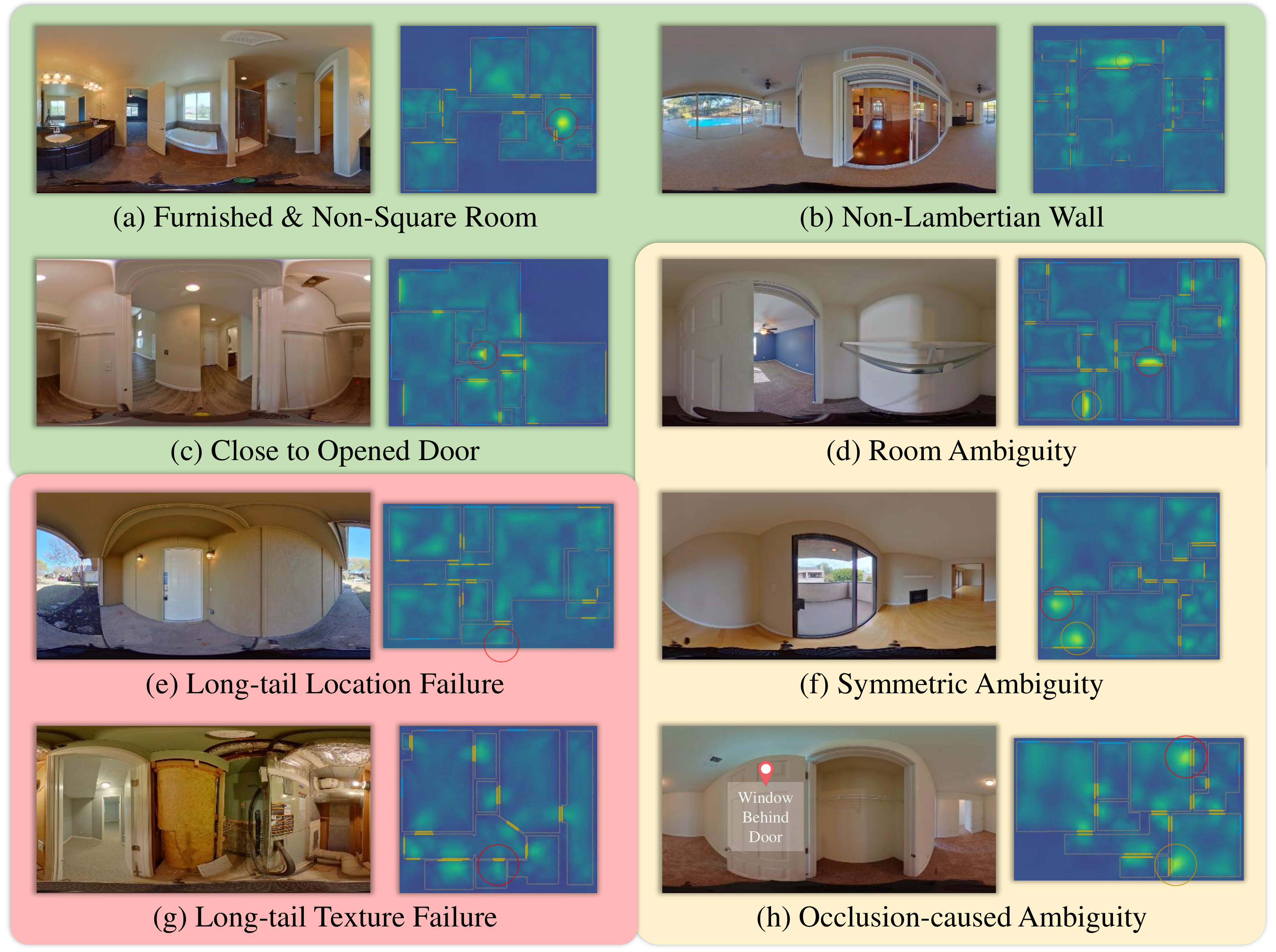}
    \vspace{-1.5em}
    \caption{\textbf{Qualitative study on method robustness under challenging cases.} Success, ambiguous and failure cases are placed inside green, yellow and red boxes respectively. The GT locations are circled red and ambiguities are circled yellow in the posterior maps where floor maps are overlayed.}
    \vspace{-0.5em}
    \label{fig:scoremaps_hard}
\end{figure}

\noindent\textbf{Qualitative Study.} In Figs.\ref{fig:scoremaps_pano},\ref{fig:scoremaps_pview}, we visualize posterior maps for \name and our baselines. Likelihoods are normalized w.r.t. their upper and lower bounds which vary among methods. We square our cosine distance for better visualization. 
Fig.\ref{fig:scoremaps_hard} shows \name is robust to challenging cases such as complex appearance and geometry conditions. While most failure cases are caused by ambiguities, some long-tail distributed location and texture also causes failure.

\noindent\textbf{Model Ablation.} In Table.\ref{table:ablation}, we show the ablation study over model components on ZInD dataset. Recall and median errors are reported for panorama and persp-90° input. 
%For coarse sampling grids, our refinement steps unquantizes the discrete estimations, while it similarly improves rotation accuracy by unquantizing our fixed-step rotation sampling. Translation accuracy is improved when the sampling density becomes a bottleneck. 
Our refinement step unquantizes the discrete estimations,  improving rotation accuracy from quantized initialization. Translation accuracy is similarly improved when the sampling density becomes a bottleneck. 
Context loss improves recall, and the improvements become more evident for queries with small FoV. 
Replacing the PointNet with a shared codebook across map points with same semantic labels makes \name agnostic to the input map domain, but slightly degrades all metrics. Omission of codebook rendering or circular feature encoding,  largely degrades all performance metrics.

\begin{figure*}[t]
    \centering
    \begin{tabular}{cc}
    \subfloat[\scriptsize{Performance over FoV. (Left: Recall) (Right: Accuracy) }] 
    {\label{fig:perf_over_fov}
    \includegraphics[width=0.48\columnwidth]{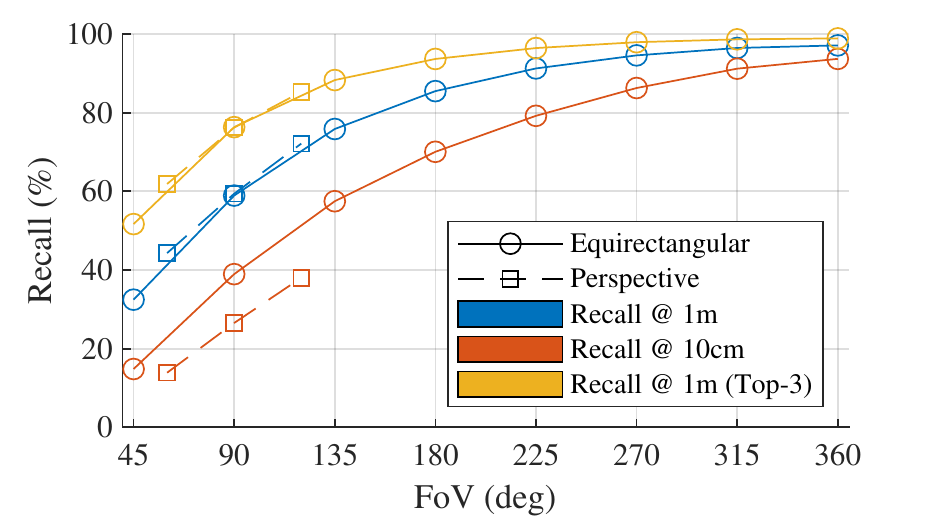} 
    \includegraphics[width=0.48\columnwidth]{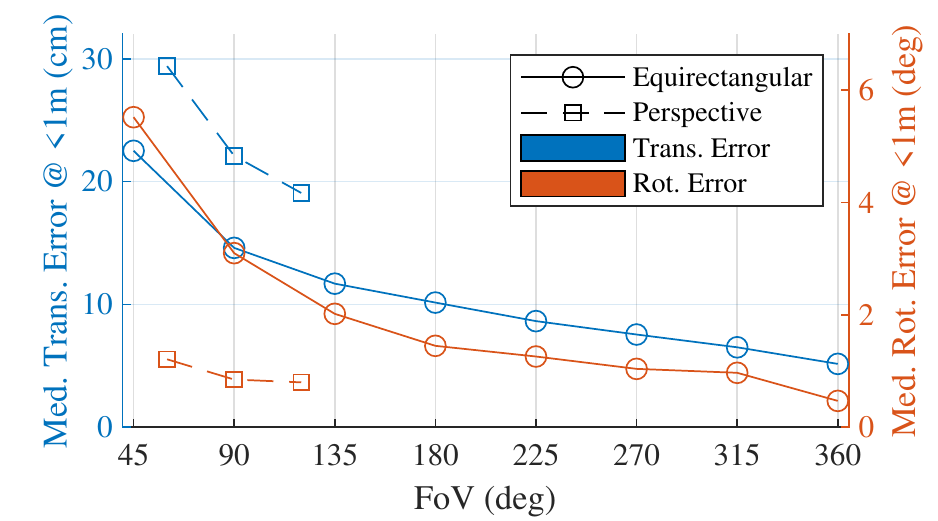} } & 
    \subfloat[\scriptsize{Performance over codebooks. (Left: Translation) (Right: Rotation)}]
    {\label{fig:perf_over_codebook}
    \includegraphics[width=0.48\columnwidth]{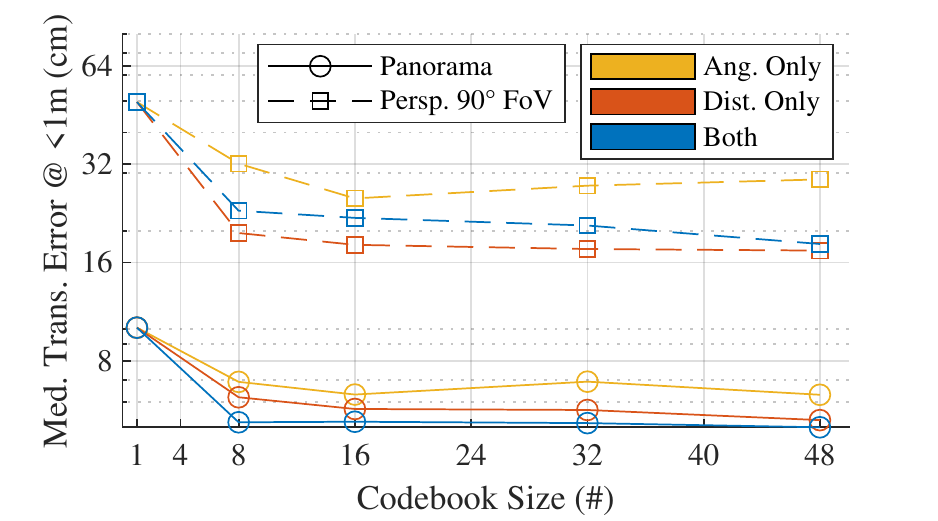}
    \includegraphics[width=0.48\columnwidth]{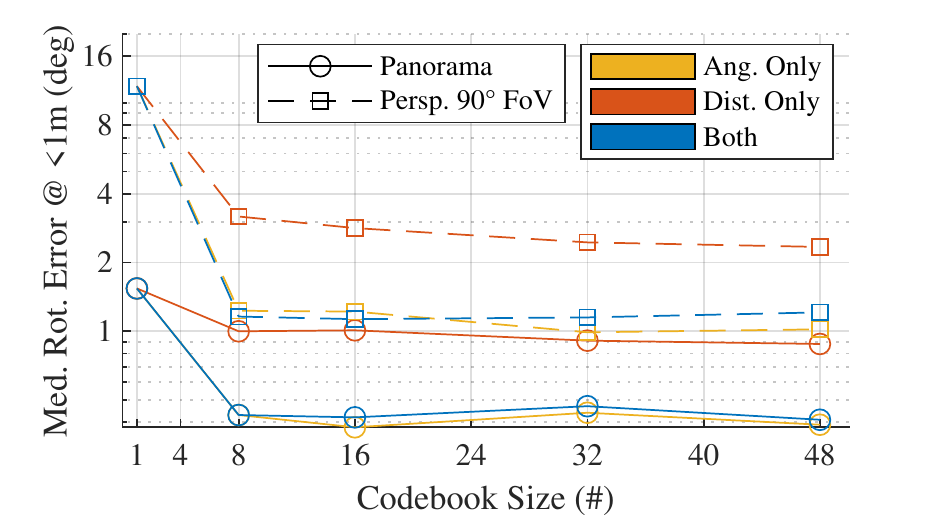}} \\
    \subfloat[\scriptsize{Performance over sampling density. (Left: Translation) (Right: Rotation)}]
    {\label{fig:perf_over_sampling_density}
    \includegraphics[width=0.48\columnwidth]{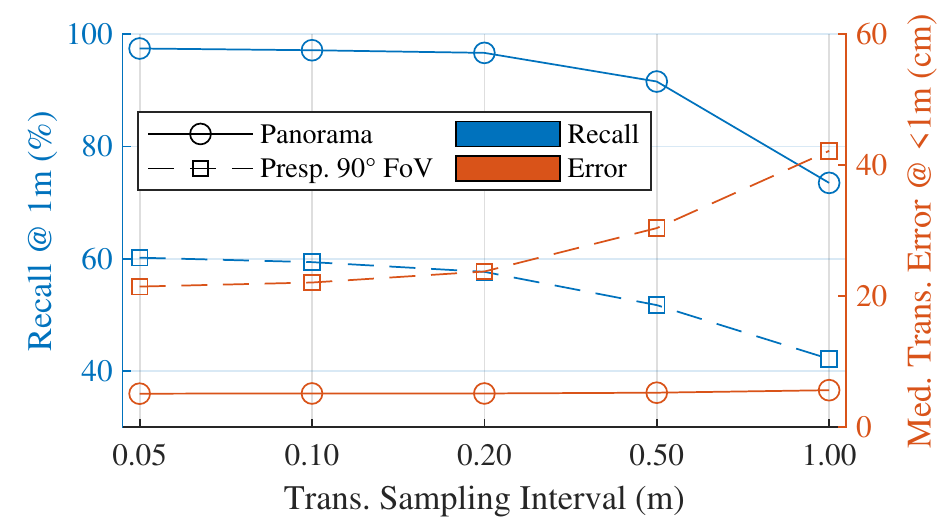}
    \includegraphics[width=0.48\columnwidth]{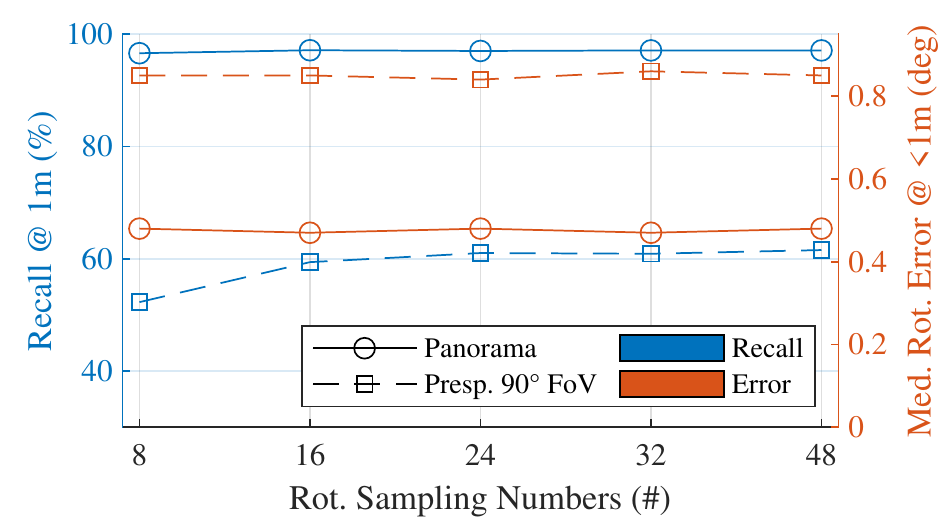}} &  
    \subfloat[\scriptsize{Performance over rendering resolution. (Left: Recall) (Right: Accuracy)}]
    {\label{fig:perf_over_rendering}
    \includegraphics[width=0.48\columnwidth]{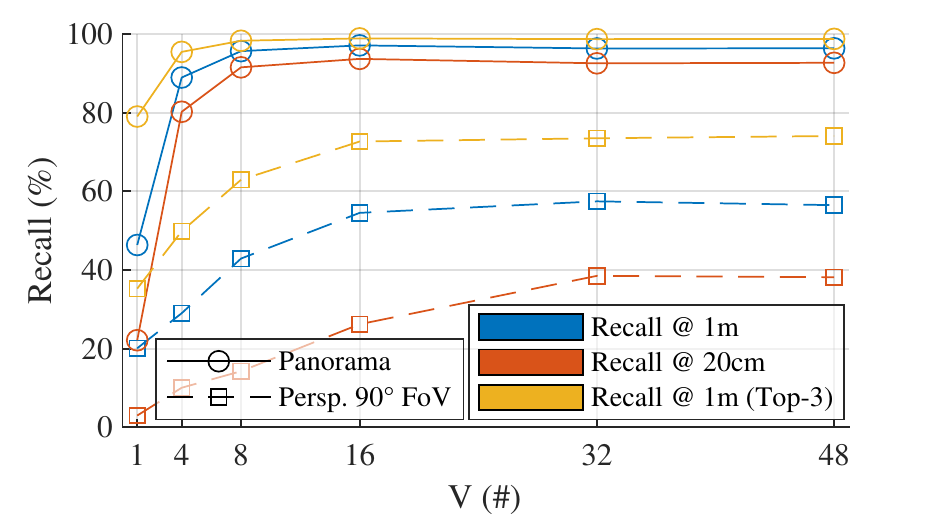}
    \includegraphics[width=0.48\columnwidth]{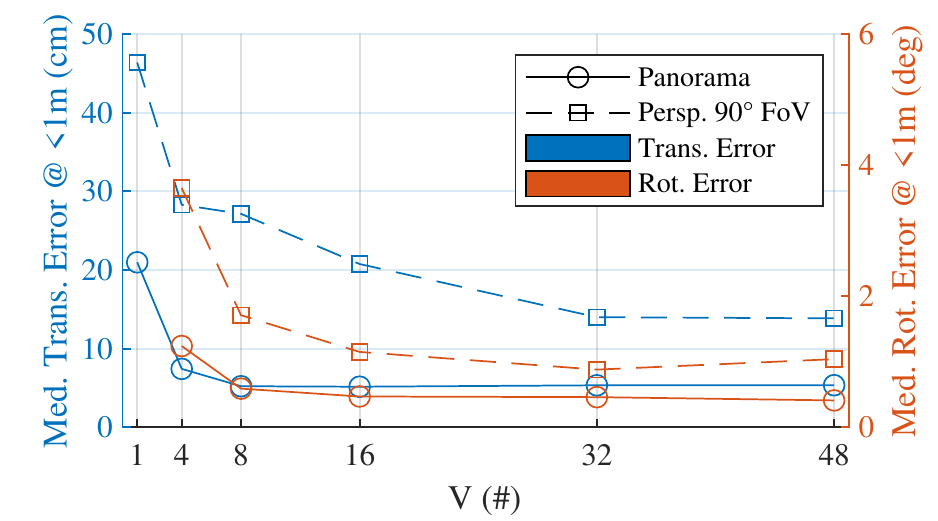}}
    \end{tabular}
    \vspace{-0.5em}
    \caption{\textbf{Detailed performance analysis.} Different query types are shown in line-types. Different metrics/configurations are shown in colors. Note that some figures have two y-axes showing different metrics in separate colors. }
    \vspace{-0.5em}
    \label{fig:perf_over_four}
\end{figure*}

\begin{figure}[]
    \centering
    \includegraphics[width=0.9\columnwidth]{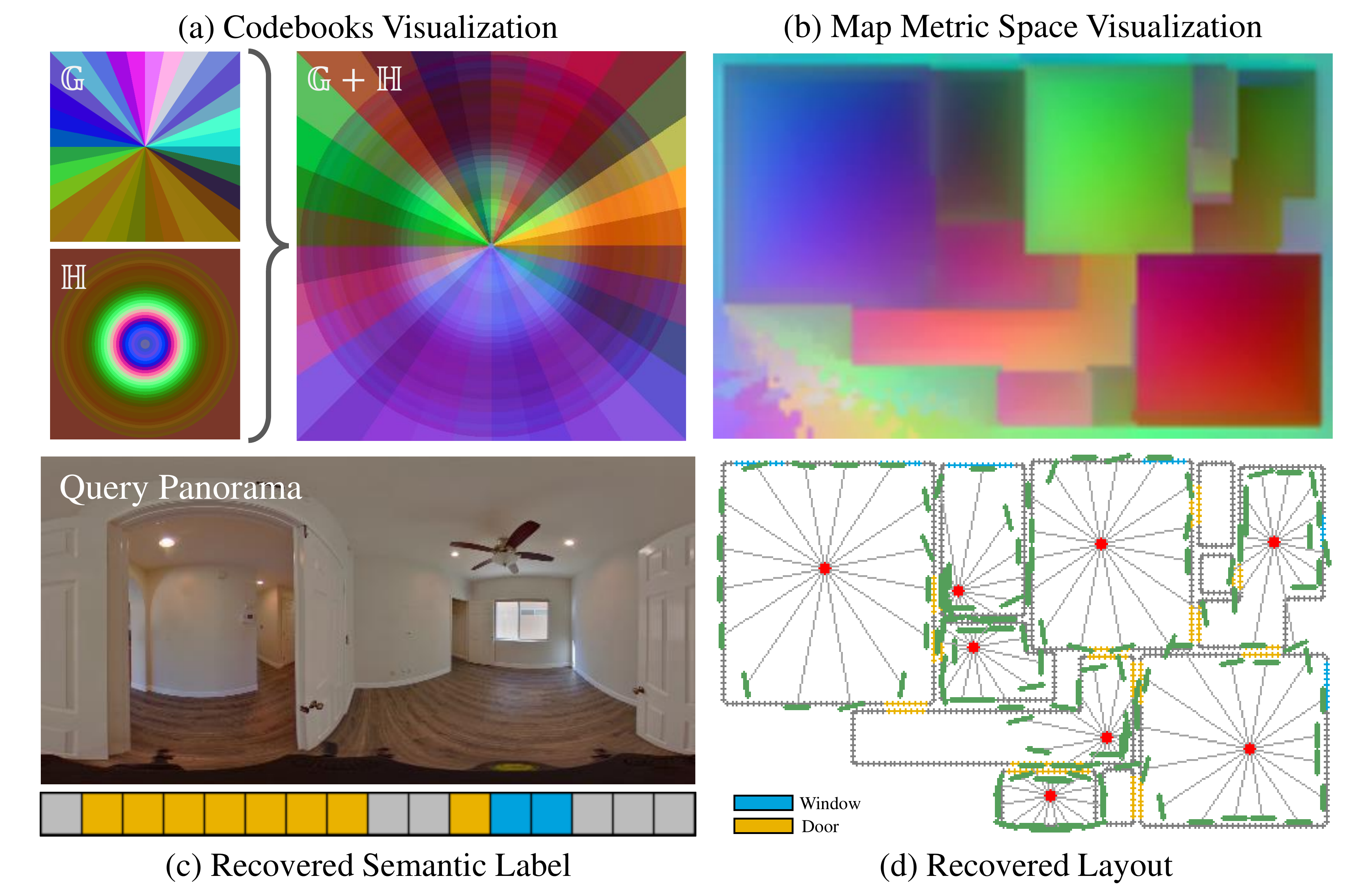}
    \vspace{-0.5em}
    \caption{\textbf{Visualization and interpretability.} (a) Codebook visualization using cosine kernel PCA. (b) Visualization of map metric space using cosine kernel t-SNE. (c,d) Recovered image semantic labels and room layout by inversely matching image circular feature exhaustively against  all map point codebooks. (d) The recovered viewing rays and incident-angles are shown in gray and green lines respectively. }
    \vspace{-1em}
    \label{fig:insight}
\end{figure}

\noindent\textbf{Performance over FoV.} In Fig.\ref{fig:perf_over_fov}, we show performance over different query image FoVs. The perspective query images have same horizontal and vertical FoV, while equirectangular query images always have 180° vertical FoV. With increasing image FoV, \name consistently gains better performance. \name has lower translation error with equirectangular views, but lower rotation error from perspective views, while their 1m-recall is similar.

\noindent\textbf{Performance over Hyper-Parameters.} Fig.\ref{fig:perf_over_codebook} shows that the incident-angle codebook improves rotation accuracy while the distance codebook is better in translation. Combining both gives the best. Performance is not sensitive to codebook size, where a modest number (i.e. 16) is satisfactory. Fig.\ref{fig:perf_over_sampling_density} shows increasing the translation sample density improves the recall and translation accuracy, but such improvement becomes marginal after 0.1m$\times$0.1m. While  recall is not sensitive to rotation sample density, where a modest number (i.e. 16) is satisfactory. Fig.\ref{fig:perf_over_rendering} shows a modest rendering resolution $V$ (i.e. 16/32) is satisfactory.

\noindent\textbf{Codebook Visualization and Interpretability.} As shown in Fig.\ref{fig:insight}(a,b), both codebook and sampled feature map exhibit smooth transitions between adjacent codes while there are distinctiveness for distant codes and different rooms. The distance codebook has less variation after a certain distance, which is a cue for choosing $d_{max}$. Furthermore, we exhaustively match image feature to the codebooks of all map points. We greedily record the code with closest distance, where the semantic label can be recovered from its map point as shown in Fig.\ref{fig:insight}(c). The distance and incident-angle can be also recovered w.r.t. to its codebook as shown in Fig.\ref{fig:insight}(d). This shows our model is implicitly learning to extract semantic and layout from the image.

\noindent\textbf{Timing.} As shown in Table.\ref{table:timing}, \name is significantly faster than existing works both in sampling and query, which allows its application in time-sensitive re-localization. Within the query, the refinement step is relatively slower since the sequential rendering is not massively parallelizable.

\section{Conclusions and Discussions}
\vspace{-0.5em}
%In this work, we proposed a latent space rendering scheme that explicitly renders features vectors within the latent space. Combining with a proposed geometrically-structured metric learning framework, we showed its efficient and accurate application in the 2D visual localization task. 
%In this work, we introduced 
\name introduces the concept of feature codebook which leverages the redundancy of overspecified feature predictions to achieve runtime dynamic latent feature assignments, while obviating additional expensive encoding processes. Moreover, we instantiate our codebook scheme into a novel latent space rendering process, where rendering dynamics are efficiently encoded into latent feature representations. 
In practice, latent space rendering enables our proposed  geometrically-structured metric learning framework to achieve state-of-art efficiency and accuracy within the 2D visual localization task.  

Besides the success that feature codebook achieved in the 2D visual localization task, the codebook scheme can also be applied in general learning tasks to replace the encoder network. In addition to the speed-up, the codebook scheme enables the seamless debugging using the (nearest-neighbor) inverse matching as shown in Fig.\ref{fig:insight}(c,d). In practice, the nearest-neighbor matching can be easily extended to extract probabilistic estimation. We believe such tool has strong potential to benefit general learning tasks in the aspects of interpretability, probabilistic estimation/calibration, and improving speed/latency.
%The proposed  geometrically-structured metric learning framework showed its efficient and accurate application in the 2D visual localization task. 

Limitations to address in future work include modeling object-based (e.g. furniture) partial map occlusions. Moreover, we have not considered the semantics between visibility and environmental state (e.g. leveraging a door's open/close status to reason about the content visibility). Finally, the scope of our latent space rendering, may be extended to 3D space or consider more complex rendering-time dynamics such as surface BRDF.

\begin{table}
    \renewcommand{\arraystretch}{1.1}
    \centering
    \resizebox{1\columnwidth}{!}{%
        \input{tabels/table_timing}
    }
    \caption{\textbf{Timing.} Performance on S3D with a single NVIDIA Tesla V100 following the algorithm configurations as in \S\ref{sec:benchmark}. }
    \vspace{-1em}
    \label{table:timing}
\end{table}

%\section{Guidance for Supplementary}
%\noindent {\bf Guidance for Supplementary Material}.
%The following contents are included in the supplemental due to space limitations:  1) Comparing uncertainty estimation vs real error distribution. 2) Robustness over noisy/sparse map. 3) Robustness over pitch/roll angle noise in perspective images 4) Matching maps to single images (i.e. global map retrieval). 5) Performance over data attributions (i.e. furnishing-level, saliency) 6) Additional implementation details. 7) Additional experiment justifications.

%%%%%%%%% REFERENCES
{\small
\bibliographystyle{ieee_fullname}
\bibliography{egbib}
}

\end{document}

%% file: tabels/table_baselines.tex
\begin{tabular}{ccccccccc|ccccccc} 
\hline
 &  & \multicolumn{7}{c|}{ZInD} & \multicolumn{7}{c}{Structured3D \, (Furnishing-Level : Full)} \\ 
\hline
Query & Method & \begin{tabular}[c]{@{}c@{}}\textless{}1m med\vspace{-0.3em}\\terr (cm)\end{tabular} & \begin{tabular}[c]{@{}c@{}}\textless{}1m med\vspace{-0.3em}\\rerr (deg)\end{tabular} & \begin{tabular}[c]{@{}c@{}}10cm recall\vspace{-0.3em}\\(\%)\end{tabular} & \begin{tabular}[c]{@{}c@{}}50cm recall\vspace{-0.3em}\\(\%)\end{tabular} & \begin{tabular}[c]{@{}c@{}}1m recall\vspace{-0.3em}\\(\%)\end{tabular} & \begin{tabular}[c]{@{}c@{}}1m \& 30°\vspace{-0.3em}\\recall (\%)\end{tabular} & \begin{tabular}[c]{@{}c@{}}top-3\vspace{-0.3em}\\1m recall (\%)\end{tabular} & \begin{tabular}[c]{@{}c@{}}\textless{}1m med\vspace{-0.3em}\\terr (cm)\end{tabular} & \begin{tabular}[c]{@{}c@{}}\textless{}1m med\vspace{-0.3em}\\rerr (deg)\end{tabular} & \begin{tabular}[c]{@{}c@{}}10cm recall\vspace{-0.3em}\\(\%)\end{tabular} & \begin{tabular}[c]{@{}c@{}}50cm recall\vspace{-0.3em}\\(\%)\end{tabular} & \begin{tabular}[c]{@{}c@{}}1m recall\vspace{-0.3em}\\(\%)\end{tabular} & \begin{tabular}[c]{@{}c@{}}1m \& 30°\vspace{-0.3em}\\recall (\%)\end{tabular} & \begin{tabular}[c]{@{}c@{}}top-3\vspace{-0.3em}\\1m recall (\%)\end{tabular} \\ 
\hline
- & Random & (70.71) & (90.00) & 0.00 & 0.61 & 2.15 & 0.26 & 5.71 & (70.71) & (90.00) & 0.00 & 0.53 & 2.36 & 0.29 & 7.48 \\ 
\hline
\multirow{5}{*}{Panorama} & PfNet & 48.77 & 15.20 & 0.70 & 19.21 & 37.15 & 28.82 & 50.58 & 44.37 & 14.97 & 1.65 & 27.52 & 47.38 & 36.48 & 64.05 \\
 & LaLaLoc & 10.65 & - & 35.61 & 71.69 & 76.00 & - & 91.62 & 6.83 & - & 58.57 & 85.98 & 87.51 & - & 98.23 \\
 & MCL & 11.88 & 5.60 & 38.96 & 86.33 & 90.15 & 85.21 & 98.66 & 6.44 & 7.18 & 57.22 & 77.49 & 86.51 & 67.12 & \textbf{99.41} \\
 & Ours \footnotesize{(no sem)} & 5.66 & 0.49 & 67.52 & 86.81 & 88.48 & 85.24 & 96.17 & 4.83 & 0.28 & 59.99 & 75.19 & 83.50 & 67.00 & 97.17 \\
 & Ours & \textbf{5.16} & \textbf{0.47} & \textbf{78.83} & \textbf{96.83} & \textbf{97.12} & \textbf{96.99} & \textbf{98.90} & \textbf{3.87} & \textbf{0.23} & \textbf{79.20} & \textbf{95.05} & \textbf{95.52} & \textbf{94.76} & 98.41 \\ 
\hline
\multirow{4}{*}{\begin{tabular}[c]{@{}c@{}}Perspective\\60° FoV\end{tabular}} & PfNet & 63.31 & 23.48 & 0.21 & 5.23 & 15.86 & 9.19 & 24.87 & 61.27 & 17.60 & 0.35 & 6.01 & 16.91 & 11.20 & 26.75 \\
 & MCL & \textbf{21.72} & 5.40 & \textbf{6.05} & 18.66 & 23.58 & 20.13 & 39.76 & 21.79 & 7.50 & 5.48 & 14.50 & 18.80 & 12.79 & 35.12 \\
 & Ours \footnotesize{(no sem)} & 26.22 & 1.27 & 3.56 & 19.40 & 24.94 & 20.95 & 42.46 & 32.95 & 1.53 & 1.89 & 16.97 & 23.69 & 14.67 & 46.38 \\
 & Ours & 29.39 & \textbf{1.21} & 4.61 & \textbf{34.42} & \textbf{44.30} & \textbf{42.10} & \textbf{61.88} & \textbf{16.97} & \textbf{0.98} & \textbf{11.02} & \textbf{41.48} & \textbf{45.96} & \textbf{43.13} & \textbf{65.65} \\ 
\hline
\multirow{4}{*}{\begin{tabular}[c]{@{}c@{}}Perspective\\90° FoV\end{tabular}} & PfNet & 56.55 & 14.24 & 0.55 & 10.97 & 26.78 & 19.63 & 38.62 & 60.00 & 12.49 & 0.59 & 8.78 & 24.34 & 18.74 & 34.41 \\
 & MCL & \textbf{17.00} & 5.08 & \textbf{11.28} & 30.18 & 34.72 & 31.41 & 53.43 & 15.41 & 6.14 & 9.55 & 22.63 & 26.81 & 20.33 & 46.91 \\
 & Ours \footnotesize{(no sem)} & 19.29 & 0.94 & 7.99 & 29.87 & 35.15 & 31.57 & 54.37 & 25.11 & 0.87 & 3.83 & 22.04 & 27.58 & 18.74 & 51.27 \\
 & Ours & 22.09 & \textbf{0.85} & 9.42 & \textbf{51.62} & \textbf{59.40} & \textbf{57.77} & \textbf{76.23} & \textbf{12.99} & \textbf{0.64} & \textbf{20.86} & \textbf{53.51} & \textbf{56.28} & \textbf{54.21} & \textbf{76.31} \\ 
\hline
\multirow{4}{*}{\begin{tabular}[c]{@{}c@{}}Perspective\\120° FoV\end{tabular}} & PfNet & 53.81 & 11.74 & 0.74 & 14.21 & 31.42 & 25.94 & 43.83 & 60.27 & 12.51 & 0.77 & 10.14 & 28.05 & 22.39 & 38.95 \\
 & MCL & \textbf{14.72} & 5.17 & \textbf{17.12} & 43.19 & 48.17 & 43.96 & 66.52 & 12.66 & 6.61 & 16.44 & 34.00 & 39.13 & 29.40 & 59.46 \\
 & Ours \footnotesize{(no sem)} & 15.56 & 0.92 & 14.19 & 42.49 & 46.99 & 43.75 & 65.86 & 22.96 & 1.03 & 7.72 & 32.76 & 40.19 & 27.64 & 63.64 \\
 & Ours & 19.07 & \textbf{0.80} & 15.18 & \textbf{65.37} & \textbf{72.14} & \textbf{71.08} & \textbf{85.28} & \textbf{11.31} & \textbf{0.70} & \textbf{30.88} & \textbf{68.83} & \textbf{70.95} & \textbf{69.77} & \textbf{87.98} \\
\hline
\end{tabular}

%% file: tabels/table_ablation.tex
\begin{tabular}{cccc|ccc} 
\hline
 & \multicolumn{3}{c|}{Panorama} & \multicolumn{3}{c}{Persp. 90° FoV} \\ 
\hline
Model & \begin{tabular}[c]{@{}c@{}}\textless{}1m med\vspace{-0.2em}\\terr~(cm)\end{tabular} & \begin{tabular}[c]{@{}c@{}}\textless{}1m med\vspace{-0.2em}\\rerr~(deg)\end{tabular} & \begin{tabular}[c]{@{}c@{}}1m recall\vspace{-0.2em}\\(\%)\end{tabular} & \begin{tabular}[c]{@{}c@{}}\textless{}1m med\vspace{-0.2em}\\terr~(cm)\end{tabular} & \begin{tabular}[c]{@{}c@{}}\textless{}1m med\vspace{-0.2em}\\rerr~(deg)\end{tabular} & \begin{tabular}[c]{@{}c@{}}1m recall\vspace{-0.2em}\\(\%)\end{tabular} \\ 
\hline
base & 7.22 & 5.45 & 97.06 & \textbf{20.59} & 5.46 & \textbf{54.55} \\
+ refine & \textbf{5.16} & \textbf{0.47} & \textbf{97.12} & 20.76 & \textbf{1.15} & 54.53 \\
- cxtloss & 8.54 & 5.45 & 95.76 & 21.57 & 5.46 & 50.52 \\
- pointnet & 7.84 & 5.46 & 95.87 & 23.46 & 5.68 & 50.26 \\
- codebook & 22.74 & 5.87 & 64.26 & 61.05 & 23.70 & 10.34 \\
- circ-feat. & 17.30 & - & 58.24 & 46.31 & - & 20.11 \\
\hline
\multicolumn{7}{r}{\footnotesize `$+$' with component \; `$-$' without component}
\end{tabular}

%% file: tabels/table_timing.tex
\begin{tabular}{ccccccc} 
\cline{1-4}\cline{6-7}
Method & Sampling Fps & Sampling Time (s) & Query Fps & ~ ~ ~ & Query Module & Time (ms) \\ 
\cline{1-4}\cline{6-7}
PfNet & 2471 $\pm$ 1328 & 48.95 $\pm$ 38.95 & 5.06 $\pm$ 1.77 &  & ResNet50 & 10.5 $\pm$ 1.2 \\
LaLaLoc & 15.47 $\pm$ 5.96 & 25.13 $\pm$ 23.67 & 0.30 $\pm$ 0.06 &  & Measure & 17.7 $\pm$ 7.4 \\
Ours & 13238 $\pm$ 1890 & 0.97 $\pm$ 1.09 & 8.31 $\pm$ 0.64 &  & Refine & 92.7 $\pm$ 16.1 \\
\cline{1-4}\cline{6-7}
\end{tabular}